\documentclass{article}

\usepackage{arxiv}

\usepackage[utf8]{inputenc} 
\usepackage[T1]{fontenc}    
\usepackage{lmodern}        
\usepackage{hyperref}       
\usepackage{url}            
\usepackage{booktabs}       
\usepackage{amsfonts}       
\usepackage{nicefrac}       
\usepackage{microtype}      
\usepackage{lipsum}
\usepackage{graphicx}

\title{A review of approaches to modeling applied vehicle routing problems}

\author{
    Konstantin Sidorov
   \\
    Adeptik \\
   \\
  \texttt{\href{mailto:ksidorov@adeptik.com}{\nolinkurl{ksidorov@adeptik.com}}} \\
   \And
    Alexander Morozov
   \\
    Adeptik \\
   \\
  \texttt{\href{mailto:morozov@adeptik.com}{\nolinkurl{morozov@adeptik.com}}} \\
  }

\usepackage{color}
\usepackage{fancyvrb}

\DefineVerbatimEnvironment{Highlighting}{Verbatim}{commandchars=\\\{\}}
\usepackage{framed}
\definecolor{shadecolor}{RGB}{248,248,248}
\newenvironment{Shaded}{\begin{snugshade}}{\end{snugshade}}

\newcommand{\KeywordTok}[1]{\textcolor[rgb]{0.13,0.29,0.53}{\textbf{#1}}}
\newcommand{\NormalTok}[1]{#1}

\newcommand{\OtherTok}[1]{\textcolor[rgb]{0.56,0.35,0.01}{#1}}

\newcommand{\StringTok}[1]{\textcolor[rgb]{0.31,0.60,0.02}{#1}}

\usepackage{amsmath}
\usepackage{amsfonts}
\usepackage{amssymb}
\usepackage{amsthm}

\usepackage{mathtools}
\usepackage{bm}

\usepackage{booktabs}
\usepackage{tabularx}
\usepackage{makecell}
\usepackage{multirow}

\usepackage[usenames,dvipsnames]{xcolor}

\usepackage{tikz}
\usetikzlibrary{mindmap}

\usepackage{setspace}
\usepackage[numbers]{natbib}

\begin{document}
\maketitle

\def\tightlist{}

\begin{abstract}
Due to the practical importance of vehicle routing problems (VRP), there exists the
ever-growing body of research in algorithms and (meta)heuristics for solving
such problems. However, the diversity of VRP domains creates the separate problem
of modeling such problems -- describing the domain entities (and, in particular, the
planning decisions), the set of valid planning decisions, and the preferences between
different plans.
In this paper, we review the approaches for modeling vehicle routing problems.
To make the comparison more straightforward, we formulate several criteria for
evaluating modeling methods reflecting the practical requirements of the development
of optimization algorithms for such problems. Finally, as a result of this comparison,
we discuss several future research avenues in the field of modeling VRP domains.
\end{abstract}

\hypertarget{introduction}{%
\section{Introduction}\label{introduction}}

Vehicle routing problems (VRP) are computational problems that can be generally formulated by the following description: given a set of transportation requests and a fleet of vehicles, determine a set of vehicle routes to perform transportation requests with the given vehicle fleet at minimum cost. \citep{Irnich2014}

It is important to note that vehicle routing problems differ in some (or all) of the following points:

\begin{itemize}
\tightlist
\item
  What are the transportation requests?
\item
  How the transportation requests must (or should) be executed?
\item
  What is a vehicle and how can it move between locations?
\item
  Which sets of routes are feasible?
\item
  How is the cost of the route set defined?
\end{itemize}

Vehicle routing problems, while being a generic and loosely defined class of problems, turn out to be of large practical importance -- it is well-known \citep{Rodrigue2020} that transportation normally has a significant contribution to the final price of the product, and the computer-aided planning can yield tangible savings to a producing company \citep{Hasle2007}. However, since VRP contain the traveling salesman problem, any VRP can be considered an NP-hard problem.

Despite being a computationally difficult problem, many different variations of VRP and their solution techniques have been extensively studied \citep{Eksioglu2009, Vidal2013, Braekers2016}. However, this amount of research creates another problem: papers on VRP frequently lack the common language for defining the problem in question (which we will further refer to as \emph{modeling language}). In particular, this creates a fragmentation not only in problem definitions but for their solution techniques.

The goal of this study is to review the existing general approaches to modeling vehicle routing problems with their strengths and weaknesses. Since VRP is a vast scope of problems, we will focus on \emph{deterministic vehicle routing problems} -- i.e., we will assume that all necessary problem information is known up-front, without having to model probability distributions (if problem information is stochastic) or tree search (if problem information is revealed sequentially, which is known as \emph{dynamic VRP} \citep{Pillac2013}).

The rest of the paper is structured as follows:

\begin{itemize}
\tightlist
\item
  Section \ref{background-vehicle-routing-problems} provides a general overview of vehicle routing problems. Since domain modeling normally facilitates the development of optimization algorithms, this section overviews not only the VRP formulations but some of the common approaches for solving them.
\item
  In Section \ref{evaluation-criteria} we introduce several criteria useful for comparing different VRP modeling approaches. This list of criteria is derived from the typical needs of developing algorithms described in the previous section.
\item
  We review the ad-hoc approaches for modeling VRP instances and the software tools aiding this kind of modeling in Section \ref{ad-hoc-vrp-models}.
\item
  In Section \ref{mathematical-programming-models} we review the mathematical programming models. Since these models also provide a way for \emph{solving} the problem instances, we also review techniques used for developing optimization algorithms for mathematical programming models corresponding to VRP instances.
\item
  We review the methods for storing the information about problem instances in a machine-readable way in Section \ref{formal-languages-for-vrp-problem-instances}. While this approach does not give much in a way of developing algorithms, it is the most ``specialized'' for VRP instances in general and transferable between different VRP problem definitions.
\item
  In Section \ref{conclusions-and-future-work} we provide the concluding remarks on the direct comparison of presented approaches.
\end{itemize}

\hypertarget{background-vehicle-routing-problems}{%
\section{Background: vehicle routing problems}\label{background-vehicle-routing-problems}}

\begin{figure}
\centering

\begin{tikzpicture}[
  mindmap, grow cyclic,
  every node/.style=concept,
  every node/.append style={execute at begin node=\hskip0pt},
  concept color=orange!40,
    level 1/.append style={
      sibling angle=90,
      font=\normalsize,
    minimum size=2.5cm,
    text width=2.25cm,
      level distance=4cm,
  },
    level 2/.append style={
      sibling angle=75,
      font=\normalsize,
    minimum size=2.35cm,
    text width=2.1cm,
    level distance=3.5cm,
    },
    level 3/.append style={
      sibling angle=70,
    font=\small,
    minimum size=2.25cm,
    text width=2cm,
    level distance=3.25cm
    },
    level 4/.append style={
      sibling angle=45,
    font=\small,
    minimum size=2.25cm,
    text width=2cm,
    level distance=3.25cm
    },
]

\node[minimum size=2.5cm,text width=2.25cm]{VRP algorithms}
child [concept color=blue!30, level distance=3.4cm] { node {Mathematical programming}
    child { node {Constraint programming}}
    child { node {Mixed-integer programming}
      child { node [align=center] {Flow \\ formulations}}
      child { node [align=center] {Set \\ partitioning formulations}}
    }
}
child [concept color=purple!50] { node {Heuristic algorithms}
    child [concept color=purple!40] { node {Constructive heuristics}
      child [
        concept color=purple!25,
      sibling angle=40,
      minimum size=1.5cm,
      text width=1.5cm,
    ] { node {\dots} }
      child { node {Nearest neighbor}}
      child { node {Savings algorithm}}
    }
    child [concept color=purple!60] { node {Metaheuristics}
      child [
        concept color=purple!25,
      sibling angle=40,
      minimum size=1.5cm,
      text width=1.5cm,
    ] { node {\dots} }
      child { node {Local search}
        child { node {Large neighborhood search}}
        child { node {Simulated annealing}}
        child { node {Tabu search}}
      }
      child { node {Evolutionary algorithms}
        child { node {Genetic algorithms}}
        child { node {Ant colony optimization}}
        child { node {Particle swarm optimization}}
      }
    }
}
child [concept color=teal!40, level distance=3.4cm] { node {Machine learning approaches}
    child { node {End to end learning}}
    child { node {Learning to configure}}
    child { node [align=center] {ML \\ alongside \\ algorithms}}
};

\end{tikzpicture}

\caption{Mind map of optimization algorithms for vehicle rounting problems.}
\label{fig:algo-mind-map}
\end{figure}
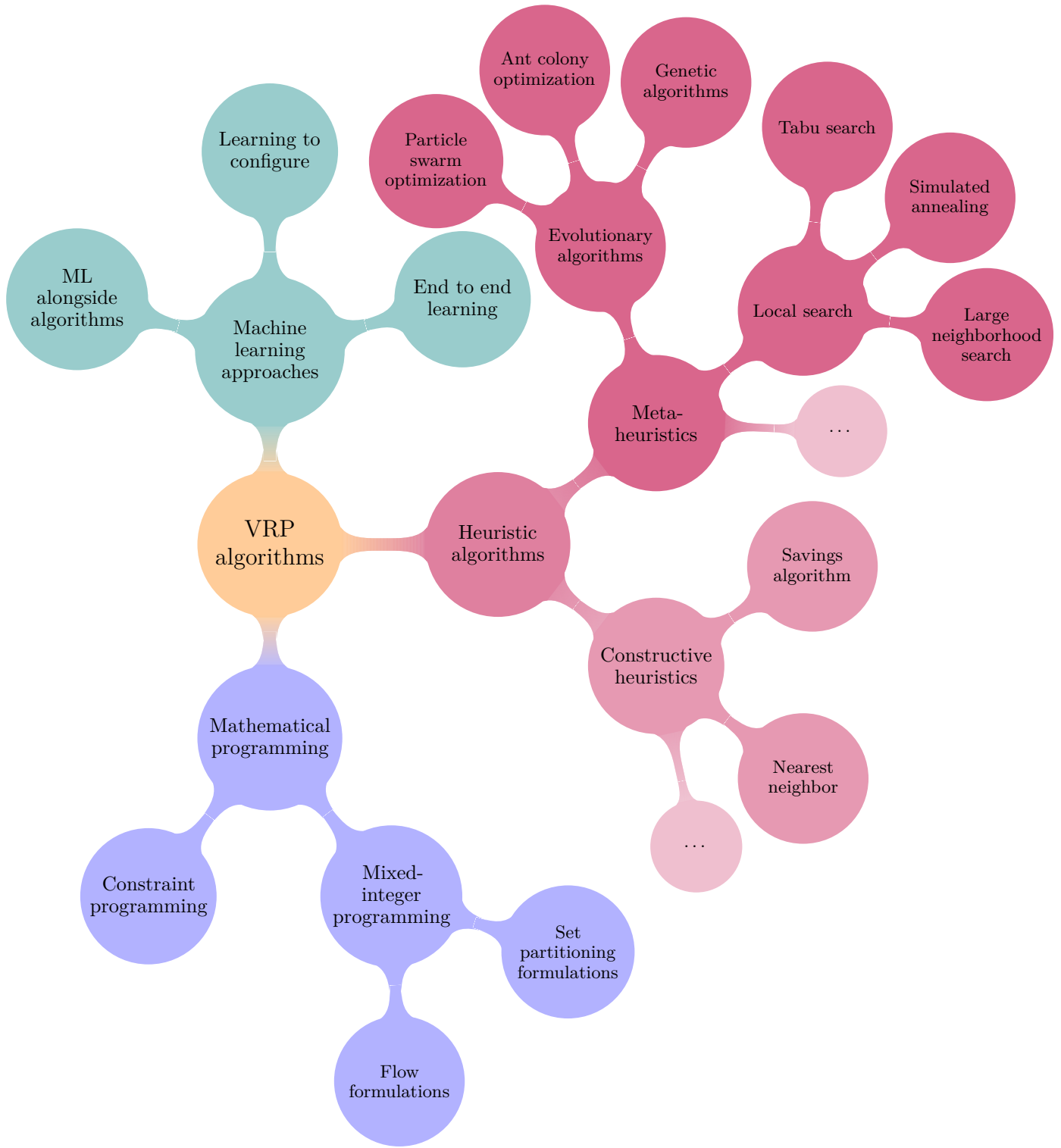

The first non-trivial VRP problem variation that has been formulated and computationally studied is \emph{traveling salesman problem}, which is the problem of finding the shortest route through a graph that visits each node exactly once. This problem has been first studied by Karl Menger, who also pointed out that the nearest-neighbor heuristic does not give the optimal route \citep{schrijver2005}.

The study of TSP has given rise to several important optimization techniques:

\begin{itemize}
\item
  \emph{Cutting plane methods} -- an algorithm for solving integer linear programs (which, in particular, contain TSP as a sub-class of problems) that refine a set of feasible values by introducing new inequalities into the problem (also known as \emph{cuts}). These methods have been introduced as an efficient method to solve non-trivial TSP instances \emph{to optimality} by George Dantzig, Delbert Ray Fulkerson, and Selmer M. Johnson in \citep{Dantzig1954}.
\item
  \emph{Branch and bound algorithms} -- a paradigm for designing optimization algorithms that prescripts to represent the solution space as a search tree and exhaustively explore its branches, \emph{discarding those that cannot improve the solution}. The discarding is normally done by maintaining lower and upper \emph{bounds} on the optimal solution for all branches and comparing them with the ``running optimal'' solution. As for the term ``branch and bound'' itself, it has been first used by Little et al.~in \citep{little1963}.
\end{itemize}

The systematic study of vehicle routing problems originates from the seminal paper \citep{dantzig1959} which introduced the real-world optimization of gasoline delivery to gas stations. The optimization problem introduced there possessed many traits typical for modern VRP domains:

\begin{itemize}
\item
  it directly involved the concept of \emph{capacity} -- on the one hand, the customers had some fixed resource demand, on the other hand, the customers that could be served by a single vehicle was constrained by the vehicle capacity;
\item
  it included both the routing component (akin to TSP) and the assignment component (the customers had to be partitioned into multiple routes).
\end{itemize}

The algorithms for solving vehicle routing problems are summarized in Figure \ref{fig:algo-mind-map} -- the remainder of this section will provide a short review of the mentioned algorithms and relevant publications.

The \emph{heuristic algorithms} are the most common algorithms for such problems and can be roughly classified as \emph{constructive heuristics} and \emph{metaheuristics}.

The constructive heuristics are procedures that construct the valid VRP solution, normally in time polynomial in the input size and without any guarantees in terms of solution quality. The most natural and general constructive heuristic is \emph{nearest neighbor algorithm}, which constructs vehicle routes incrementally by choosing the nearest available location on every step, starting from the depot. The first efficient construction heuristic for such problems has been proposed in \citep{clarke1964} and is known as \emph{savings algorithm}. The algorithm proceeds as following:

\begin{enumerate}
\def\labelenumi{\arabic{enumi}.}
\item
  Make a route for each of the nodes \(v_k\) that visits only that node: \(v_0 \to v_k \to v_0\), where \(v_0\) is the depot.
\item
  For each pair of customers \(v_i, v_j\) compute savings for merging this pair as \(s_{ij} = d_{i0}+d_{j0} -d_{ij}\).
\item
  For each pair of customers \(v_i, v_j\) \emph{in decreasing order of savings} merge the corresponding routes if:

  \begin{itemize}
  \item
    customers \(v_i, v_j\) are not already on the same route,
  \item
    both customers are directly connected to the depot (i.e., are either first or last in the route)
  \item
    the merge does not violate the capacity constraint.
  \end{itemize}
\item
  Return the resulting solution after no more merges can be applied.
\end{enumerate}

While such procedures are normally guaranteed to terminate quickly and return a valid solution, in general, such a solution can be arbitrarily suboptimal. For this reason, the most efficient VRP algorithms rely on some kind of search procedure. The most common approach for VRP domains is \emph{metaheuristics} -- general procedures that can be used for designing an approximation algorithm for a problem at hand.

The simplest and most common example of a metaheuristic is \emph{local search} -- a family of search procedures that have the following common traits:

\begin{itemize}
\tightlist
\item
  they start the search from a single valid solution (normally generated by some constructive heuristic),
\item
  on each iteration, they produce a \emph{neighborhood} of the incumbent solution -- a set of valid solutions, normally of polynomial size and consisting of solutions ``similar'' to the incumbent,
\item
  for a neighborhood of the incumbent, they choose one solution and accept it (promote it to the incumbent solution) according to some rule -- which can be both deterministic (accept if better than incumbent) and stochastic (accept if better, otherwise accept with some probability).
\end{itemize}

For a general review of local search applications in VRP, see \citep{braysy2005}.

While this scheme can be implemented naïvely by choosing some appropriate neighborhood and using the trivial acceptance rule, varying the components of local search gives rise to even more efficient metaheuristics. For example, the \emph{simulated annealing} is a metaheuristic that closely follows this scheme, differing only in the acceptance rule: if the incumbent solution has the objective value \(J\), and the new solution has the objective \(J'\), then the new solution is accepted with the probability\[
\begin{cases}   1 & J' > J \\  \exp\left(\frac{J' - J}{T}\right) & \text{otherwise,}\end{cases}
\]

where \(T > 0\) is a \emph{temperature} -- an algorithm parameter controlling the ``randomness'' of transitions that degrade the solutions. In particular,

\begin{itemize}
\item
  as \(T \to 0\), the transition rule degenerates into a ``greedy rule'' that only accepts improving solutions;
\item
  on the other hand, as \(T \to +\infty\), the transition rule degenerates into a ``random search'' which accepts \emph{any} proposed solution regardless of its quality.
\end{itemize}

These two extremes are normally used in the following way: the temperature is initialized to some large value at the start of the algorithm and gradually reduced according to some \emph{annealing schedule}, thus interpolating between the random search in the beginning and ``greedy'' search on later stages of the algorithm.

The first usage of the term ``simulated annealing'', as well as the first application of this technique to TSP, can be found at \citep{Kirkpatrick671}. The first study that used simulated annealing for solving VRP is \citep{osman1993}; the paper \citep{chiang1996} has the first description of non-hybrid simulation annealing strategy for VRP.

Another local search technique widely known in the VRP community is \emph{tabu search}. The metaheuristic can be generally described by changing the acceptance rule: in tabu search, the new solution can be accepted even if it is worse than the incumbent, but the search algorithm maintains the \emph{tabu list} -- a special data structures that maintain the recent solutions and forbid returning to them.

The first application of tabu search to VRP can be found at \citep{gendreau1994}; for the more comprehensive review of tabu search algorithms and domain-specific techniques used for implementing such algorithms, see \citep{glover1997}.

Another important class of algorithms is \emph{evolutionary algorithms} -- unlike local search, they maintain a \emph{population} of valid solutions and iteratively improve the population as a whole. The prime example is the family of \emph{genetic algorithms}, which repeats the following operations with the population:

\begin{enumerate}
\def\labelenumi{\arabic{enumi}.}
\tightlist
\item
  \emph{Selection}. This operation removes the part of existing solutions, normally preferring to remove worse solutions.
\item
  \emph{Crossover}. This operation selects multiple pairs of solutions (typically in a way that prefers the best solutions) and constructs new solutions from them\footnote{This process can be informally seen as breeding a new generation of solutions.}.
\item
  \emph{Mutation.} At this stage, the solutions of the population are \emph{individually} and randomly changed in a way similar to a single step of local search.
\end{enumerate}

The first work that described the genetic algorithm competitive with the tabu search on vehicle routing problems is \citep{baker2003}. For the classic introduction to the field of genetic algorithms, see \citep{10.5555/534133}.

A comprehensive treatment of metaheuristic algorithms, their implementation details, and theoretical analysis, can be found in the monograph \citep{handbook2010}.

The \emph{mathematical programming} models are mathematical models constructed by the following scheme:

\begin{itemize}
\item
  The model introduces \emph{decision variables} for domain decisions (i.e., assignment of customers to vehicles or a sequence of visits for vehicles).
\item
  The set of valid assignments for decision variables is defined by the set of \emph{constraints} -- predicates involving decision variables, normally having some restricted form.
\item
  The assignments for decision variables are ranked by \emph{objective function} -- a function that maps such assignments to real values. The objective function is used to encode preferences between valid solutions.
\end{itemize}

In the context of VRP domains, the most relevant classes of mathematical programming models are:

\begin{itemize}
\tightlist
\item
  \emph{Mixed-integer programming} (MIP), which is a class of mathematical programming problems in which the problem domain is specified by inequalities over real and/or integer variables. A very important special case is mixed-integer \emph{linear} programming, which further restricts the problem to linear inequalities and linear objective function.
\item
  \emph{Constraint programming} (CP) is another class of mathematical programming models, in which the decision variables are restricted to \emph{finite sets}, but the constraints can be specified by \emph{arbitrary relations} between decision variables.
\end{itemize}

For further discussion of mathematical programming models for VRP domains, see Section \ref{mathematical-programming-models}.

Recent developments in the field of machine learning have created the \emph{machine learning} approach to solving vehicle routing problems. The general overview of the machine learning applications to the more general context of combinatorial optimization can be found in the review \citep{BENGIO2021405}. Such approaches can be classified by their place in the resulting algorithm:

\begin{itemize}
\tightlist
\item
  \emph{End to end learning} approaches attempt to build a model that \emph{directly} solves the problem instance. This approach is implemented in many papers utilizing such an approach -- some of the examples are \citep{bello2016neural}, \citep{dai2017learning}, \citep{nazari2018reinforcement}, and \citep{kool2018attention}.
\item
  \emph{Learning to configure algorithms} approaches, on the contrary, don't influence the search process directly -- rather, they attempt to provide useful information for other solving approaches (e.g., by selecting the algorithm parameters). This approach is similar to \emph{meta-learning} approach in machine learning, which is used to select the most performing machine learning model for a given dataset \citep{vanschoren2019}. Applications of this approach to vehicle routing problems include \citep{tyasnurita2017} and \citep{gutierrez-rodriguez2019}, while paper \citep{kanda2016} investigates feature selection for applying this approach to TSP.
\item
  \emph{Machine learning alongside optimization algorithms} approaches combine the previous two approaches in the sense that they build a model for guiding the search algorithm. The research in this area for VRP is more sparse than for the other two approaches -- one of the recent papers that studies this approach for VRP domains is \citep{gao2020learn}.
\end{itemize}

\hypertarget{evaluation-criteria}{%
\section{Evaluation criteria}\label{evaluation-criteria}}

The wide range of VRP models and their solution methods creates a separate problem for \emph{modeling} VRP domains. In the remainder of this review we will introduce several modeling methods -- but, since the field of vehicle routing problems is diverse, we need a way to define the strong and weak sides of such methods. To this end, we have to define a set of criteria that will be used to evaluate modeling approaches. From the discussion of the approaches to solving VRP domains, we derived the following criteria:

\begin{itemize}
\tightlist
\item
  \emph{Expressiveness of modeling language}. The modeling languages satisfying this criterion should provide modeling constructs for various situations typical to VRP domains (which include temporal constraints or modeling different entity types).
\item
  \emph{Unambiguous domain entity modeling}. This criterion separates the modeling approaches that inherently allow modeling standard VRP constructs. In particular, this criterion is somewhat similar to the following quote of \emph{Zen of Python} \citep{pep20}: ``There should be one --- and preferably only one --- obvious way to do it''.
\item
  \emph{Separation of domain-level and instance-level modeling}. Using this criterion, we would like to separate the modeling systems that have a clear divide between the part that models the domain and the part that encodes the actual problem instance data. In the VRP context, this is particularly useful, because, as we have already seen, modeling the domain entities in VRP domains can itself be a non-trivial task.
\item
  \emph{Additional complexity of developing optimization algorithms for problem domains}. This criterion will be used to show how much overhead does the modeling approach introduces to the task of algorithm development (in comparison with the ad-hoc modeling).
\item
  \emph{Portability of optimization modules across problem domains}. This criterion separates the systems for which developing the optimization algorithms working \emph{across} the problem domains is not substantially harder than developing the optimization algorithms for a \emph{single} problem domain.
\end{itemize}

\hypertarget{ad-hoc-vrp-models}{%
\section{Ad-hoc VRP models}\label{ad-hoc-vrp-models}}

The most natural way to model the optimization problem in question is by laying out its domain model in code -- an approach which we will further refer to as \emph{ad-hoc modeling}.

As we will see further, there are more ``principled'' approaches to VRP modeling -- however, the strong side of ad-hoc modeling is that it scales well if there is \emph{only one} problem domain that needs to be dealt with. In particular, this approach is dominant for \emph{rich VRPs}, which are commonly defined as VRP models with multiple heterogeneous entities and/or constraints. A taxonomy for such problems, as well as a statistical definition for ``richness'', is provided in \citep{lahyani2015}. In particular, even though \emph{VRP-REP}, the collection of VRP domains and their respective instances \citep{Mendoza2014}, provides its standard for defining instances, at the moment of writing \emph{less than a half} of domains represented at VRP-REP, actually use this standard. (For further information about VRP-REP, see Section \ref{embedding-into-general-purpose-languages}.)

One of the most well-known software toolkits for this modeling approach is OptaPlanner \citep{optaplanner} -- a Java library branded as ``an AI constraint solver'' which aids the development of optimization algorithms by providing the workflow with the following steps:

\begin{itemize}
\tightlist
\item
  \emph{Domain model introduction}. At this stage, the entities of the domain are defined as POJOs\footnote{\emph{Plain Old Java Objects} -- a software engineering concept that normally denotes a Java object. This term is normally understood as an antonym for objects used in more complicated frameworks that, in particular, frequently have to implement prespecified interfaces, extend prespecified classes or contain prespecified annotations \citep{pojo}.} in the same way it is normally done in software engineering.
\item
  \emph{Search space definition}. At this stage, the developer annotates some of the entities as \emph{planning} entities (a concept also known as ``decision variables'') and defines their respective ranges. This is normally done by defining a POJO for a solution (e.g., in VRP case it would likely contain a mapping between vehicles and their routes) and specifying constraints (and an objective as well) with a factory class that returns a list of hard/soft constraints for given domain entities.
\item
  \emph{Finding the solution}. In the case of OptaPlanner, this stage reduces to the invocation of the built-in \texttt{Solver} class which contains the actual optimization logic. It should be stressed, however, that the OptaPlanner only provides building blocks for implementing local search algorithms, somewhat limiting the performance of the resulting algorithm.
\end{itemize}

\hypertarget{mathematical-programming-models}{%
\section{Mathematical programming models}\label{mathematical-programming-models}}

The more principled approach to modeling VRP problems is to state them as \emph{mathematical programming} problems, i.e.~representing the problem domain in question as a problem of selecting an element of a set (normally given by a list of \emph{constraints} parameterized by \emph{decision variables}) that maximizes/minimizes the \emph{objective function} (a function that maps the elements of the domain to real numbers).

This approach is well known far beyond VRP context -- in fact, it has a vast body of research both for theoretical properties of such problems and for its practical applications. Since the scope of this work is limited to the VRP applications, we will introduce only a (small) part of mathematical programming research directly relevant to the VRP context. More exactly, we will review two classes of models mentioned in Section \ref{background-vehicle-routing-problems}: mixed-integer programming (Section \ref{mixed-integer-programming-models}) and constraint programming (Section \ref{constraint-programming-models}).

\hypertarget{mixed-integer-programming-models}{%
\subsection{Mixed-integer programming models}\label{mixed-integer-programming-models}}

Aside from ad-hoc modeling, the most common approach to model routing problems is \emph{mixed-integer programming} -- a class of mathematical optimization models that employ numeric variables, some types of algebraic constraints (e.g., linear or quadratic), and integrality constraints on some of the variables.

The simplest VRP variant that can be modeled with this approach is \emph{traveling salesman problem}. Without loss of generality we will assume that the graph has \(n\) nodes \(1, 2, \dotsc, n\) and edges having lengths \(c_{ij}\).

The common issue of TSP modeling via MIP is the \emph{subtour elimination}, i.e.~imposing a constraint that forbids selecting two or more different tours partitioning the node set and otherwise giving a feasible solution. There are two common formulations for the TSP which differ by the way they define the subtour elimination constraints:

\begin{itemize}
\tightlist
\item
  \emph{Miller-Tucker-Zemlin formulation} (MTZ) defines a variable \(x_{ij} \in { 0, 1 }\) for each edge and a dummy variable \(u_i \in \mathbb{Z}, 1 \le u_i \le n - 1\) for each node (except the node \(1\), which is implicitly assumed to start the tour). The subtour elimination is implemented by introducing a constraint \(u_i - u_j + nx_{ij} \le n-1\) for each edge \((i, j)\) not incident to node \(1\) -- thus, the dummy variables implement the tour ordering. It can be shown that this constraint (1) successfully eliminates subtour solutions and (2) remains feasible for any complete tour defined by \(x\)-variables. \citep{Miller1960}
\item
  \emph{Dantzig--Fulkerson--Johnson formulation} (DFJ) defines only variables \(x_{ij} \in { 0, 1 }\) for each edge \((i, j)\) and implements the subtour elimination by imposing a flow constraint for each node subset \(Q \subset \{1, \dotsc, n\}\) such that \(1 < \#Q < n\) (i.e., it has at least two nodes but is a strict subset): \(\sum_{i \in Q, j \in Q, i \neq j} x_{ij} \ge 1\). \citep{Dantzig1954}
\end{itemize}

While MTZ formulation, unlike DFJ, can be formulated with a \emph{polynomial-size} integer linear program, it is known \citep{Velednitsky2017} that DFJ formulation ends up being stronger in a sense that the relaxations resulting from it give tighter upper bounds on the optimal solution value.

Unlike TSP, however, VRP has an additional decision-making level, since the customers should not only be served but also partitioned between vehicles. Two common approaches to this are \emph{flow formulations} and \emph{set partitioning formulations}.

Flow formulations are the direct generalizations of DFJ formulation, which introduce a binary variable for each edge and constrain their value in a way that preserves both the flow structure of the tours and does not generate sub-tours.

The typical example of such formulation has been first introduced by \citep{Laporte1986}. Let \(G=(V, E, \mathbf{c})\) be a weighted graph for which we have to solve the \emph{capacitated VRP} -- a problem of selecting \(K\) routes of shortest total length partitioning the set of customers and starting from the depot node \(0 \in V\). The model in question introduces the decision variable \(x(e) \in \{ 0, 1\}\) for each edge of the network that represents the decision on whether to take the edge in the route. Let \(\delta^+(v)\) and \(\delta^-(v)\) be respectively a set of edges outgoing from \(v\) and incoming to \(v\), and for any set of edges \(E' \subset E\) we will introduce the notation \(x(E') := \sum_{e\in E'}x(e)\). In this framework, the final MIP is as follows:

\begin{align*}
  & \bm{c}^T \bm{x} \to \min \\
  \forall v \in V &: x(\delta^+(v)) = 1 \\
  \forall v \in V &: x(\delta^-(v)) = 1 \\
  & x(\delta^+(0)) = K \\
  & \mathbf{1}^T \bm{\lambda} = K \\
  \forall V' \subset V &: x(\delta^+(V')) \ge r(V')
\end{align*}

The constraints model (respectively) the degree restrictions (each node should be entered and left exactly once), the route count, and the sub-tour elimination constraints.

This type of MIP models, while simple, has several major drawbacks:

\begin{itemize}
\tightlist
\item
  Flow models are constrained to the case when the objective and constraints are linear with respect to the variables. In particular, this makes the modeling of non-linear constructs, such as time windows or commodity compatibility, practically impossible.
\item
  Flow models do not give any way to point the vehicle that traverses any given edge. In particular, this kind of model cannot be applied for modeling problems in which the travel cost for an arc/feasibility of the vehicle move depends on the vehicle that travels through it.
\end{itemize}

More advanced approach to modeling VRP is set partition formulations. The key difference from flow formulations is that the variables are introduced not for the edges, but for the \emph{circuits} -- cyclic routes that go through the customers starting from the depot. For capacitated VRP the first such model has been introduced in \citep{Balinski1964}. Let \(\Lambda = \{ \Lambda_1, \dotsc, \Lambda_C \}\) be the set of all circuits, and for all vertices \(v\) let \(\bm{a}_v \in \{0, 1\}^C\) be the vector that specifies which circuits pass through this vertex: \(a_{vc} = 1 \Leftrightarrow v \in \Lambda_c\). The model in question introduces the decision variables \(\bm{\lambda} \in \{0, 1\}^C\) and formulates the following MIP:

\begin{align*}
  & \bm{c}^T \bm{\lambda} \to \min \\
  \forall v \in V &: \bm{a}_v^T \bm{\lambda} = 1 \\
  & \bm{1}^T \bm{\lambda} = K \\
  \forall r &: \lambda_r \in \{ 0, 1 \}
\end{align*}

The decision variables \(\bm{\lambda}\) correspond to all feasible circuits, and the constraints require that (respectively) each customer has to be served exactly once, a total of \(K\) circuits has to be selected and each circuit can be selected at most once.

The way to address this issue is \emph{column generation}, which can be described for VRP as follows:

\begin{enumerate}
\def\labelenumi{\arabic{enumi}.}
\tightlist
\item
  Choose a (small) set of circuits for optimization.
\item
  Solve the linear relaxation of the VRP formulation and derive the optimal dual variables for constraints.
\item
  Find a circuit that (a) has not been selected yet and (b) has the most negative reduced cost. The optimization problem solved at this step is called the \emph{pricing problem}.
\item
  If no such circuit exists, return the best incumbent solution (it is optimal). If not, add the new circuit to the set of circuits and return to (2).
\end{enumerate}

Set partition formulations, unlike flow formulations, resolve the feasibility issue \emph{implicitly}, i.e.~by specifying the set of routes rather than by direct constructions of constraints. This makes such an approach much more general than the flow formulations. However, its key problem is the number of variables, which is at least exponential with respect to the instance size.

While MIP is a common approach for mathematical optimization, it has several limitations which make it unsuitable for many practical problems:

\begin{itemize}
\tightlist
\item
  \emph{Modeling constructs}. MIP solvers normally are very restricted in the form of constraints -- i.e., linear programs only allow the usage of linear expressions. However, many practical constraints are inherently non-linear -- for instance, modeling different vehicle types, compatibility between resources or depot constraints, while technically viable to MIP modeling, do not result in models that are suitable for modern MIP solvers.
\item
  \emph{Non-trivial solution methods}. Unlike most typical MIP use cases, modeling VRP with MIP requires more than just constructing a model for a MIP solver -- it also requires a careful approach for orchestrating MIP solver runs (column generation in set partitioning formulations, valid inequality generation in the flow models). This point, in particular, weakens this approach with respect to ad-hoc modeling, for which a vast family of metaheuristic strategies has been devised.
\item
  \emph{Temporal constraints}. The major problem with employing time windows in the MIP models is that the resulting models are inherently nonlinear -- for example, see the \citep{Baker1983} for the description of the first such model -- and if they can be successfully linearized, the resulting models normally contain many big-M constraints, which are known to be computationally difficult for MIP solvers.
\end{itemize}

\hypertarget{constraint-programming-models}{%
\subsection{Constraint programming models}\label{constraint-programming-models}}

Another approach to VRP modeling is rewriting the problem in question as a \emph{constraint programming} problem -- an optimization problem with the following traits:

\begin{itemize}
\tightlist
\item
  The variables are specified by their domain, which is given as a \emph{finite} set.
\item
  The set of feasible solutions is defined by the set of constraints -- unlike MIP constraints, CP constraints are specified as an \emph{arbitrary} relation specified on a subset of variables.
\end{itemize}

Constraint programming is known to be a feasible approach for domains in which the optimization problem (or satisfiability problem) is specified on a set without a ``good'' geometry, like type inference \citep{Chandra2016} or automated planning \citep{Garrido2009}.

The most common approach to modeling VRP problems as CP problems is as follows:

\begin{enumerate}
\def\labelenumi{\arabic{enumi}.}
\tightlist
\item
  For each customer model the next customer visited by the same vehicle as a decision variable.
\item
  Introduce any other domain quantities (current time, load, etc.) as decision variables, and define them for each trip using the path constraint.
\item
  Introduce the domain constraints by appropriately constraining the decision variables introduced in (2).
\end{enumerate}

One of the first studies to explore this approach is \citep{DeBacker2000}, which introduces a CP model for a VRP variant with vehicle capacities and time windows. This study sidesteps the traditional tree search algorithms, introducing instead of them ad-hoc local search algorithms. This can be seen as a way of exploiting the domain knowledge (in this case, the local search operators known to work well on VRP problems) in the constraint programming framework.

Constraint programming can also be used to optimize parts of the search algorithm. For example, \citep{Shaw1998} presents an approach that uses LNS for searching the solution space, and constraint programming to efficiently restore the parts of the solution during the construction phase.

While this approach is less studied than MIP modeling, its key practical benefit is the \emph{rich modeling language} it provides. Unlike MIP models which are heavily restricted in the expressions that can be used in them, CP modeling languages allow using a wide range of constructs, normally including arbitrary logical expressions and custom constructs such as ``all-different''. The prime example of such modeling language is MiniZinc \citep{Nethercote2007}, which directly supports several constructs directly useful for modeling vehicle routing problems, such as circuit constraints, path constraints, or ``all-different'' constraint, which are impossible to model in the MIP framework practically.

The downsides of reducing a VRP problem to a CP problem are:

\begin{itemize}
\tightlist
\item
  \emph{Non-trivial modeling choices}. The flexibility of the modeling approach frequently introduces several ways to implement the same construct, which can subtly influence the performance of the solver. The simplest relevant example is the traveling salesman problem, for which multiple fundamentally different MiniZinc models can be devised; see \citep{Wallace2020} for a review of this topic.
\item
  \emph{Complexity of the constraint programming}. The normal downside of having a flexible modeling language is the complexity of algorithms for resolving this language. CP, in general, is an NP-complete problem -- while there are CP subcases that allow polynomial algorithms, such as Datalog, they turn out to be not sufficient for VRP modeling cases.
\end{itemize}

\hypertarget{formal-languages-for-vrp-problem-instances}{%
\section{Formal languages for VRP problem instances}\label{formal-languages-for-vrp-problem-instances}}

While the amount of VRP research grows, one of the problems common for many research papers is the benchmark problem set -- in most cases, it is defined in an ad-hoc way and specified with the instance statistics and their distribution (e.g., ``the depot is located in a point uniformly distributed at the unit square''). The existence of such benchmark set (as seen with, for example, MIPLIB \citep{Gleixner2021} and MiniZinc Challenge \citep{Stuckey2010}) can be helpful in several ways, such as:

\begin{itemize}
\tightlist
\item
  the unified, reproducible benchmark for further research,
\item
  unambiguous evaluation method for new optimization techniques,
\item
  centralized location for new problem instances.
\end{itemize}

The main approaches for defining and collecting VRP instances are:

\begin{itemize}
\tightlist
\item
  extending and customizing plain-text formats for similar problems,
\item
  embedding an instance specification into a general-purpose document language.
\end{itemize}

\hypertarget{customizing-plain-text-formats}{%
\subsection{Customizing plain-text formats}\label{customizing-plain-text-formats}}

One of the most widespread formats for describing VRP instances is TSPLIB and its variations. TSPLIB \citep{Reinelt1991} is a plain-text format for traveling salesman problems (and some of its variants) which mandates the following instance structure:

\begin{itemize}
\tightlist
\item
  \emph{Problem specification part}. In this part the problem type (such as TSP or capacitated VRP) is explicitly declared; this part defines what inputs can be used further. In particular, one of the problem types explicitly defined in the specification is the standard capacitated vehicle routing problem.
\item
  \emph{Data part}. In this part, the entities from the instance (such as the depot, vehicles, and customers) are completely defined with all of their features.
\end{itemize}

The list of problems that can be modeled by TSPLIB is explicitly specified by the specification and contains the following problems:

\begin{itemize}
\item
  TSP, both in symmetrical and asymmetrical forms -- the former differing in the additional assumption that the distance from A to B is the same as the distance from B to A.
\item
  \emph{Hamiltonian cycle problem} -- a problem of testing whether a graph has a Hamiltonian cycle.
\item
  \emph{Sequential ordering problem} -- a restricted version of asymmetrical TSP with an additional set of \emph{precedence} restrictions requiring that for some locations A and B the route has to visit A before B.
\item
  \emph{Capacitated VRP}.
\end{itemize}

For example, to specify the capacitated VRP, the TSPLIB file has to provide the specification part and the data part. As a running example, we will demonstrate how to model the CVRP with 4 nodes, 2 depots, and a vehicle capacity of 100. The specification part would be structured as follows:

\begin{itemize}
\item
  The type of problem is given by the line \texttt{TYPE\ :\ CVRP}.
\item
  The number of nodes is given by the line \texttt{DIMENSION\ :\ 6} , where 6 stands for the \emph{total} number of nodes.
\item
  The vehicle capacity is given by the line \texttt{CAPACITY\ :\ 100}.
\item
  The distances between the locations can be given in different ways -- however, for brevity, we will use the geographical distance, which is specified by lines \texttt{EDGE\_WEIGHT\_TYPE\ :\ GEO} and \texttt{NODE\_COORD\_TYPE\ :\ TWOD\_COORDINATES} (the latter used for specifying the latitude and longitude).
\end{itemize}

As for the data part, it is given by:

\begin{itemize}
\item
  The node locations are given in \texttt{NODE\_COORD\_SECTION} by writing a numerical identifier and a list of coordinates for each of the locations. In our example, this might look as follows:

\begin{verbatim}
1 48.0 46.0
2 48.1 45.9
3 48.2 46.0
4 47.9 46.2
5 47.8 46.1
6 47.7 45.8
\end{verbatim}
\item
  The depot nodes are listed in section \texttt{DEPOT\_SECTION}; for example, we can specify that nodes 2 and 4 are depots as follows:

\begin{verbatim}
2
4
-1
\end{verbatim}
\item
  The demands of the remaining nodes are given in \texttt{DEMAND\_SECTION} (the depot demands also have to be filled in by zero values). In our example, this may look as follows:

\begin{verbatim}
1 50
2 0
3 70
4 0
5 30
6 60
\end{verbatim}
\end{itemize}

On this example we can already see unique traits of such representations useful for VRP modeling:

\begin{itemize}
\tightlist
\item
  TSPLIB gives a direct, unambiguous model for domain entities (albeit cluttered between file parts): this model is, in fact, explicitly stated in the specification.
\item
  TSPLIB allows switching between different domain models, even if the choices for that are rather restricted.
\item
  Last but not least, TSPLIB is \emph{tailored} for modeling VRP domains -- in particular, it contains vehicles and locations as \emph{native} constructs.
\end{itemize}

However, there are several obvious drawbacks for general-purpose vehicle routing problem modeling:

\begin{itemize}
\tightlist
\item
  \emph{Ontological complexity of the underlying problems}. Unlike MIP or CP, VRP is not a well-defined class of problems. This creates difficulties for any format that explicitly depends on the problem structure -- in fact, this problem is a key feature of VRP that makes it harder for formal modeling than MIP or CP.
\item
  \emph{Poor extensibility of the instance format}. While the TSPLIB format itself is strictly defined, the extension mechanism for it is not regulated. For this reason, the proposed extensions of the format are defined only with respect to the original format -- neither the interoperability of the extensions nor their logical consistency rules are not defined, which creates additional difficulties for any unification attempts.
\item
  \emph{Lack of distinct separation between domain model and instance data}. Arguably, the problem specification part of TSPLIB is supposed to be a domain model, but it contains multiple numeric fields that are not essential for the problem domain (for example, the number of vehicles and locations).
\end{itemize}

Some of these drawbacks are related to the fact that the TSPLIB offers an ad-hoc format for storing the problem information -- and, indeed, some of them have been successfully resolved by leveraging more general-purpose representations. One of the most successful such attempts is presented in the next subsection.

\hypertarget{embedding-into-general-purpose-languages}{%
\subsection{Embedding into general-purpose languages}\label{embedding-into-general-purpose-languages}}

Another approach to model this class of problems is using some existing language, defining its subset, and adapting existing tools for processing this language for domain requirements. The most well-known effort of this kind is VRP-REP \citep{Mendoza2014}, which is a collaborative platform that allows users to share instances with the public or privately with platform editors and referees, as well as solutions to the published instances.

VRP-REP gives an option for using the specification format -- which can be described as an XML document \citep{world2006extensible} with some predefined syntactic structure of the problem instances. It should be explicitly noted that this specification is not mandatory for uploading data to VRP-REP, and thus it also contains instances written in custom (and even ad-hoc) formats.

The latest version of the specification (0.5.0, January 8, 2016) is implemented as XML schema \citep{thompson2009w3c} -- an XML document providing the formal syntactic structure of the instances, as well as the documentation of the syntactic structures. The problem instance is provided by the element \texttt{\textless{}instance\textgreater{}} containing the following children:

\begin{itemize}
\item
  The element \texttt{\textless{}info\textgreater{}} stores instance metadata -- the dataset name and the instance name.
\item
  The element \texttt{\textless{}network\textgreater{}} describes the nodes of the network and the links between nodes, which are given by child elements \texttt{\textless{}nodes\textgreater{}} and \texttt{\textless{}links\textgreater{}} respectively:

  \begin{itemize}
  \tightlist
  \item
    The \texttt{\textless{}nodes\textgreater{}} element contains a collection of \texttt{\textless{}node\textgreater{}} elements describing individual nodes. Each of such elements contains an identifier, a type (0 for depots, 1 for customers) and coordinates. For instance, one of the customer nodes from the example above would be described as:
  \end{itemize}

\begin{Shaded}
\begin{Highlighting}[]
\NormalTok{\textless{}}\KeywordTok{node}\OtherTok{ id=}\StringTok{"1"}\OtherTok{ type=}\StringTok{"1"}\NormalTok{\textgreater{}}
\NormalTok{  \textless{}}\KeywordTok{latitude}\NormalTok{\textgreater{}48.0\textless{}/}\KeywordTok{latitude}\NormalTok{\textgreater{}}
\NormalTok{  \textless{}}\KeywordTok{longitude}\NormalTok{\textgreater{}46.0\textless{}/}\KeywordTok{longitude}\NormalTok{\textgreater{}}
\NormalTok{\textless{}/}\KeywordTok{node}\NormalTok{\textgreater{}}
\end{Highlighting}
\end{Shaded}

  \begin{itemize}
  \tightlist
  \item
    The \texttt{\textless{}links\textgreater{}} element contains a collection of \texttt{\textless{}link\textgreater{}} elements describing individual links. Each of such elements contains a head, a tail (node identifiers) and optional descriptions, such as length, travel time, time window, etc. For instance, to specify that the travel time between node 1 and 3 is 10 minutes, one can use the following construction:
  \end{itemize}

\begin{Shaded}
\begin{Highlighting}[]
\NormalTok{\textless{}}\KeywordTok{link}\OtherTok{ head=}\StringTok{"1"}\OtherTok{ tail=}\StringTok{"3"}\NormalTok{\textgreater{}}
\NormalTok{  \textless{}}\KeywordTok{travel\_time}\NormalTok{\textgreater{}10\textless{}/}\KeywordTok{travel\_time}\NormalTok{\textgreater{}}
\NormalTok{\textless{}/}\KeywordTok{link}\NormalTok{\textgreater{}}
\end{Highlighting}
\end{Shaded}
\item
  The \texttt{\textless{}fleet\textgreater{}} element describes the fleet available for executing the requests and is described by a set of vehicle profiles provided in \texttt{\textless{}vehicle\_profile\textgreater{}} elements. Each vehicle profile is defined by its departure nodes, arrival nodes, as well as with such VRP-specific characteristics as capacity, speed profile, costs, etc. For example, to specify a single vehicle with a fixed capacity that can start from any depot, one can use the following node:

\begin{Shaded}
\begin{Highlighting}[]
\NormalTok{\textless{}}\KeywordTok{vehicle\_profile}\OtherTok{ type=}\StringTok{"1"}\NormalTok{\textgreater{}}
\NormalTok{  \textless{}}\KeywordTok{departure\_node}\NormalTok{\textgreater{}2\textless{}/}\KeywordTok{departure\_node}\NormalTok{\textgreater{}}
\NormalTok{  \textless{}}\KeywordTok{departure\_node}\NormalTok{\textgreater{}4\textless{}/}\KeywordTok{departure\_node}\NormalTok{\textgreater{}}
\NormalTok{  \textless{}}\KeywordTok{arrival\_at\_any\_node}\NormalTok{ /\textgreater{}}
\NormalTok{  \textless{}}\KeywordTok{capacity}\NormalTok{\textgreater{}100\textless{}/}\KeywordTok{capacity}\NormalTok{\textgreater{}}
\NormalTok{\textless{}/}\KeywordTok{vehicle\_profile}\NormalTok{\textgreater{}}
\end{Highlighting}
\end{Shaded}
\item
  The \texttt{\textless{}requests\textgreater{}} element contains the information about the demands of the network nodes. Each request is specified by the \texttt{\textless{}request\textgreater{}} node and provides information about the requesting node, the amount, the time window, etc. The request for the first node can be modeled with the following node:

\begin{Shaded}
\begin{Highlighting}[]
\NormalTok{\textless{}}\KeywordTok{request}\OtherTok{ id=}\StringTok{"1"}\OtherTok{ node=}\StringTok{"1"}\NormalTok{\textgreater{}}
\NormalTok{  \textless{}}\KeywordTok{quantity}\NormalTok{\textgreater{}50\textless{}/}\KeywordTok{quantity}\NormalTok{\textgreater{}}
\NormalTok{\textless{}/}\KeywordTok{request}\NormalTok{\textgreater{}}
\end{Highlighting}
\end{Shaded}
\item
  Optional elements \texttt{\textless{}resources\textgreater{}} and \texttt{\textless{}drivers\textgreater{}} can be provided to model the domains in which it is important to (respectively) introduce the resources (for example, fuel or charge) and differentiate not only vehicles but their drivers (for example, to model special vehicle types requiring some special skillset or compatibility.
\end{itemize}

Using XML schema for the definition of the instance structure is an improvement to the plain-text format. Indeed, it offloads the ontological complexity of VRP to XML toolchain, and (partially) solves the extensibility problem by involving users in the discussion of the XML specification. Another strength of this approach is the multitude of modeling constructs provided by the specification, which cover various practical situations.

However, this approach also has a clear drawback in \emph{lack of clear ``problem specification''}: the problem specification is controlled by the XML schema, which allows constructing instances (which corresponds to TSPLIB ``data part''), but fails to give tools for constructing a specification of the problem at hand -- i.e., defining custom entities and the relations between them. Even though TSPLIB does not completely manage to separate these two sides of the problem, its specification introduces the explicit divide between them -- for this reason, this shortcoming of VRP-REP can be seen as a ``step backward'' with respect to existing approaches.

Another weakness of this approach is the \emph{complexity of the specification} -- to support so many various constructs, the specification has to explicitly include all of them, which results in the XML schema of 1351 lines. This can partially explain the fact that many VRP-REP users choose not to use the official specification -- moving to the new format requires understanding this specification first, which turns out to be an issue due to its sheer size.

\hypertarget{conclusions-and-future-work}{%
\section{Conclusions and future work}\label{conclusions-and-future-work}}

Given the growing economical importance of vehicle routing problems, the new approaches for modeling them develop. For this reason, it becomes increasingly more important to be aware not only of their existence but about their strengths and weaknesses. In this review we introduced the most important modeling approaches for vehicle routing problems; the results of our analysis (with respect to the criteria from Section \ref{evaluation-criteria}) are summarized in Table \ref{tab:comparison}. The main conclusions for the studied approaches are as follows:

\begin{itemize}
\tightlist
\item
  \emph{Ad-hoc modeling} is normally used to directly transfer the domain model into source code. This approach sacrifices the modeling flexibility, which, in particular, hinders the portability of optimization modules. Since the most common approaches for VRP are metaheuristic approaches, the ad-hoc approach is effective for developing optimization algorithms -- this approach leverages the toolchain of the selected programming language to aid the algorithm development.
\item
  \emph{Mathematical programming modeling} explicitly allows separating the domain model from the instance data. However, since such models are not tailored for VRP constructs, most rich models involve using unintuitive modeling constructs (and choosing between them, which also affects the algorithm quality). Using such models also involves using rather complex schemes for solving them (for example, the column generation in MIP models) -- which, on the other hand, end up being transferable between different models.
\item
  \emph{Formal language modeling} eliminates modeling ambiguity by design but does not provide a boundary between domain model and instance data. For this reason, the complexity and portability of optimization algorithms heavily depend on the language specification.
\end{itemize}

\newcolumntype{b}{>{\centering\arraybackslash}X}
\newcolumntype{s}{>{\centering\arraybackslash \hsize=0.5\hsize}X}
\newcommand{\heading}[1]{\multicolumn{1}{c}{#1}}
\def\tabularxcolumn#1{m{#1}}
\begin{table}[htbp]
\caption{Comparison of modeling approaches introduced in the review}
\label{tab:comparison}
\begin{tabularx}{\textwidth}{bsss}
\hline
Criterion & Ad-hoc modeling & Mathematical \newline programming & Formal languages \\
\hline
Expressiveness of modeling language & \textbf{Expressive} & Lacks multiple typical features of VRP domains & \textbf{Expressive} \\
Unambiguous domain entity modeling & \textbf{Unambiguous} & Ambiguous & \textbf{Unambiguous} \\
Separation of domain modeling & No separation & \textbf{Clear separation} & Minimal separation, not extensible \\
Complexity of developing optimization algorithms & \textbf{Low} & High & \textbf{Low} \\
Portability of optimization modules & Not portable & \textbf{Portable} & \textbf{Portable} \\
\hline
\end{tabularx}
\end{table}

From this comparison, it is clear that the key direction for further research is developing the modeling technology having a more clear balance between the evaluation criteria. In our opinion, the approach that is the most suitable for this development is the approach based on formal languages. In that case, the key problem is how to integrate the strengths of other approaches with the formal language in question.

\hypertarget{funding}{%
\section*{Funding}\label{funding}}
\addcontentsline{toc}{section}{Funding}

This research did not receive any specific grant from funding agencies in the public, commercial, or not-for-profit sectors.


\end{document}